\documentstyle[epsf,a4]{article}
\title{Assessing Complexity Results in Feature Theories}
\author{Marten Trautwein%
	\thanks{This research was supported by the
	Linguistic Research Foundation, which is funded by the Netherlands
	organisation for scientific research, NWO}
	}
\date{Department of Mathematics and Computer Science\\
University of Amsterdam\\
E-mail: mtrautwe@fwi.uva.nl}

\begin{document}
\maketitle
%
\newenvironment{proof}
{\noindent{\bf Proof}}%
{\noindent\framebox[1em]{\tiny Pr} \prs\vspace{3mm}\prs}
%
\newenvironment{ex}%
{\prs\vspace{3mm}\prs\noindent{\bf Example(s)}}%
{\noindent\framebox[1em]{\tiny Ex} \prs\vspace{3mm}\prs}
%
\newenvironment{desc}%
{\begin{list}{}{\setlength{\itemsep}{0mm}\setlength{\parsep}{0mm}}}%
{\end{list}}
%
\newtheorem{defin}{Definition}[section]
%
\newtheorem{thm}{Theorem}[section]
%
\newtheorem{lem}[thm]{Lemma}
%
\newtheorem{fact}[thm]{Fact}
\newcommand{\prs}{\par}
\newcommand{\avm}[1]{\left[\begin{array}{ll} #1 \end{array}\right]}
\newcommand{\fn}[1]{\mbox{\sc #1}}
\newcommand{\fv}[1]{\mbox{\em #1}}
\newcommand{\rs}[1]{\fbox{#1}}
\newcommand{\path}[1]{\langle #1 \rangle}

\begin{abstract}
In this paper, we assess the complexity results of
formalisms that describe the
feature theories used in computational linguistics.
We show that from these complexity results
no immediate conclusions can be drawn about the complexity of
the recognition
problem of unification grammars using these feature
theories.

On the one hand, the complexity of feature
theories does not provide an upper bound for the complexity
of such unification grammars.
On the other hand, the complexity of feature
theories need not provide a lower bound.
Therefore, we argue for formalisms that describe
actual unification grammars instead of feature theories.
Thus the complexity results of these formalisms
judge upon the hardness of unification
grammars in computational linguistics.
\end{abstract}

\section{Introduction}
Recently, there has been a growing interest in research on formalizing
feature theory.
Some formalisms that appeared lately are the feature algebra of
\cite{baader.ea:93}, the modal logic of \cite{blackburn.ea:93},
the deterministic finite automata of \cite{kasper.ea:90}, and
the first-order predicate logic of \cite{smolka:92}.
These formalisms describe the use of feature theory in
computational linguistics. They are a source of interesting
technical research, and various complexity results have been achieved.
However, we argue that such formalisms offer little help to
computational linguists in practice.
The grammatical theories used in computational linguistics do not
consist of bare feature theories.
The feature theories that are used in computational linguistics are
contained in unification grammars. These unification grammars
consist of constituent structure components, and feature theories.
We claim that the complexity results from the formalisms do no
longer hold when a feature theory
and a constituent structure component are combined into
a unification grammar.

In this paper, we will focus on the complexity results that are obtained
from formalizing feature theories.
We will prove that these complexity results do not hold if we consider
unification grammars that use these feature
theories in addition to a constituent structure component.
First we will show, that the complexity of
a unification grammar theory may be higher than the
complexity of its feature theory and constituent structure components.
Second we will explain, that the complexity of a unification grammar
may be lower than the complexity of the formalized feature theory.

Both proofs put the complexity results that have been achieved in a
different perspective. The first proof
implies that the complexity of a feature
theory does not provide an upper bound for the complexity
of grammars using that feature theory. The second proof
implies that the  complexity of a feature
theory might not provide a lower bound for the complexity
of grammars using that feature theory.
Therefore, we argue that if one is interested in the complexity of
unification grammars that are used in grammars, one should look at the
complexity of these unification grammars as a whole.
No insight in the complexity of a unification grammar is gained by
looking only at the complexity of its components in isolation.

The outline of this paper is as follows. The next section contains
the preliminaries on complexity theory and feature theory. In
Section~\ref{Ssft}, we introduce a simple feature theory:
a feature theory with only reentrance.
In Section~\ref{Snub}, we
present a unification grammar that uses this simple feature theory.
We show that the recognition problem of this grammar is harder than
the unification problem of the feature theory and the recognition
problem of the constituent structure component.
In Section~\ref{Snlb}, we explain why
the recognition problem of a unification grammar might be of lower
complexity than the unification problem of the feature theory.
In Section~\ref{Sconc},
we present our conclusions.

\section{Preliminaries}
\paragraph{Complexity Theory.}
In complexity theory one tries to determine the complexity of
problems. The complexity is measured by the amount of time and
space needed to solve a problem. Usually, one considers decision
problems: problems that are answered  `Yes' or `No'.
Often we are interested in the distinction between tractable
and intractable problems. A problem is tractable if
its solution requires an amount of steps that is polynomial in the
size of the input: we say that the problem requires
polynomial time. Likewise, we speak of linear time, etcetera.
The tractable problems are also called `P problems'.
The intractable problems are called `NP-hard problems'. The easiest
intractable problems are the `NP-complete problems'. It is
unknown whether NP-complete problems have polynomial time solutions.
However we know, that solutions for NP-complete problems can
be guessed and checked in polynomial time. It is strongly
believed that the class of P problems and the class of
NP-complete problems are different, although this is yet
unproven.

There is a direct manner to determine the upper bound
complexity of a problem,
if there is an algorithm that solves the problem: determine the
complexity of that algorithm.
An indirect way to determine the lower bound
complexity of a problem is the reduction.
A reduction from some problem $A$ to some problem $B$ maps
instances of problem $A$ onto instances of problem $B$.

The reductions that we will consider are known as polynomial time,
many-one reductions. These many-one reductions are subject to
two conditions:
(1) the reductions are easy to compute, and
(2) the reductions preserve the answers.
A reduction from $A$ to $B$ is easy to compute, if the mapping takes
polynomial time.
A reduction preserves answers if the answer to the instance of $A$ is
the same as the answer to the instance of $B$. That is, the answer
to the instance of $A$ is `Yes' if, and only if,
the answer to the instance of $B$ is also `Yes'.

A reduction is an elegant way to classify a problem as intractable.
Suppose problem $B$ is a problem with unknown complexity.
Let there be a reduction $f$ from an NP-hard problem $A$ to problem $B$.
Furthermore, let $f$ conform to the two conditions above.
By an indirect proof, it follows from this reduction that
$B$ is at least as hard as $A$.
Hence $B$ is also an NP-hard problem. If we also prove that we can guess
a solution for $B$ and check that guessed solution in polynomial time,
then $B$ is an NP-complete problem.

A well-known NP-complete problem is {\sc Satisfiability} (SAT).
\begin{defin}
{\sc Satisfiability} \\
{\sc Instance:}
 A formula $\varphi$, from propositional logic, in conjunctive
 normalform.  \\
{\sc Question:}
 Is there an assignment of truth-values to the propositional
 variables of $\varphi$, such that $\varphi$ is true?
\end{defin}

The instances of {\sc Satisfiability} are formulas in conjunctive
normalform, i.e., the formulas are conjunctions of
clauses. The clauses are disjunctions of literals, and the literals are
positive and negative occurrences of propositional variables.
We call formula $\varphi$ a satisfiable formula
if an assignment exists that makes formula $\varphi$ true.

An assignment assigns either the value true or
the value false to each propositional variable.
Given such an assignment, we can determine the truth-value of a
formula.
The formula $\varphi \,=\, (\gamma_1\wedge\ldots\wedge\gamma_m)$
is true if, and only if, each clause, $\gamma_i$, is true.
A clause $\gamma \,=\, (l_1 \vee\ldots\vee l_m)$
is true if, and only if, at least one literal, $l_i$, is true.
A positive (negative) literal, $l_i = p_j$ ($l_i = \overline{p_j}$),
is true if, and only if, the variable $p_j$ is
assigned the value true (false).

\paragraph{Feature theory.}
Although there is no such thing as a universal feature theory, there
is a general understanding of its abstract objects.
These abstract objects describe the internal information or
properties of words and phrases.
Properties that these abstract objects typically have
are the case, the gender, the number, and the tense of
words and phrases.

The properties of abstract objects can be combined to form new
abstract objects. This operation is called unification.
The unification of abstract objects combines all the properties of
these abstract objects, provided that the properties are
not contradictory.

All kinds of additions to these rudiments of feature theory have been
presented in the literature. We will not discuss them here, but
refer to Section~\ref{Ssft}, in which we introduce a feature
theory that serves our purposes.

\section{A simple feature theory}\label{Ssft}
In this section we will present a simple
feature theory. The feature theory contains reentrance, but
no negation or disjunction.
Although this feature theory is simple, it contains
many aspects from other feature theories.
In addition, Section~\ref{Snub} shows
that combining this simple feature theory with a simple
constituent structure component results in a
difficult unification grammar.

In the first part of this section, we will
formalize the notion of a feature theory.
In the second part of this section, we will
present an algorithm that solves
the unification problem in an amount of time that is quadratic
in the size of its input.
This part should convince the reader that the feature theory
is indeed simple.

\paragraph{The feature theory formally.}
Although a universal feature theory does not exist,
there is a general understanding of its objects.
The object of feature theories are abstract linguistic objects,
e.g., an object `sentence', an object `masculine third person singular',
an object `verb', an object `noun phrase'. These
abstract objects have properties, like, tense, number, predicate,
subject. The values of these properties are either atomic, like,
present and singular, or abstract objects, like, verb and noun
phrase.

The abstract objects can be represented as rooted graphs
(`feature-graphs'). The nodes of these graphs stand for abstract
objects, and the edges represent the properties.
More formally, a feature-graph is either a pair $(a, \emptyset)$,
where $a$ is an
atomic value and $\emptyset$ is the empty set, or a pair $(x, E)$,
where $x$ is a root node, and $E$ is a finite, possibly empty set of
edges such that (1) for each property and all nodes there is at
most one edge that represents the property departing from the node,
and (2) if there
is an edge in $E$ from node $y$ to node $z$, then there is a path
in $E$ leading from node $x$ to node $y$.

As an example consider the following abstract objects and
simplified feature-graph.
\begin{ex}
\begin{itemize}
\item {\em Sentence: A man walks} \\
	This abstract object has property \fn{tense} with value
	\fv{present}, property \fn{subject} with value {\em Noun phrase:
	A man}, and property \fn{predicate} with value {\em Verb: walks}.
\item {\em Noun phrase: A man} \\
	has property \fn{number} with value \fv{singular}.
\item {\em Verb: walks} \\
	also has property \fn{number} with value \fv{singular}.
\end{itemize}
\end{ex}

\begin{figure}[ht]
\epsffile{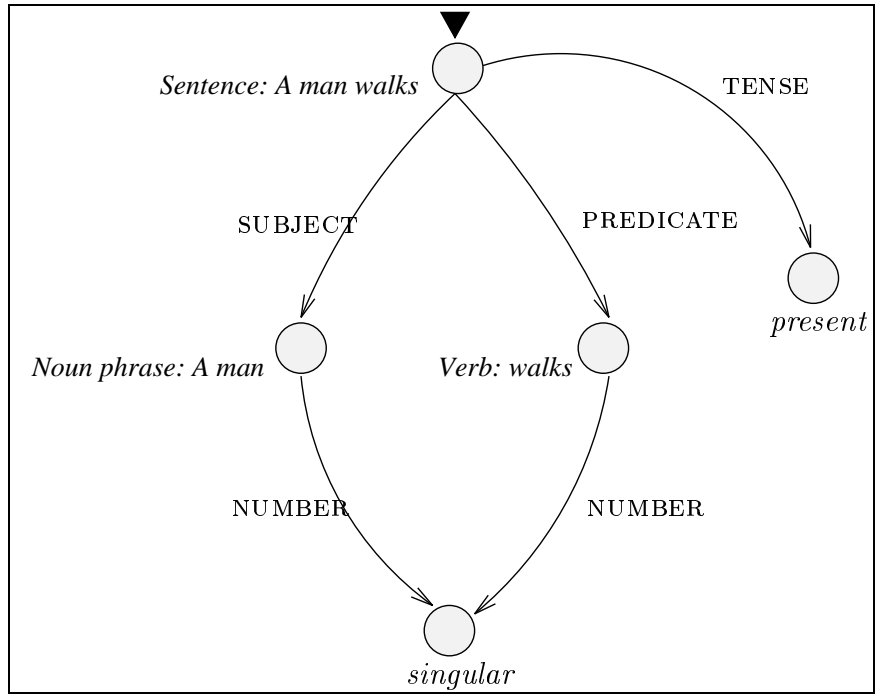}
\caption{A simplified feature-graph for `A man walks'.\label{FSman}}
\end{figure}

The abstract objects are fully described by their properties and their
values.
Multiple descriptions for the properties and values of the abstract
linguistic objects are presented in the literature.
A formal description language for these
properties and values of the abstract linguistic objects is
a sublanguage of predicate logic with equality, $F_L$,
introduced by \cite{smolka:92}.

Assume three pair-wise disjunct sets of symbols: the set of constants
$A$, the set of variables $V$, and the set of attributes $L$.
The attributes (denoted by $f, g, h$ or capitalized strings)
correspond to the properties of
the abstract objects,
the variables (denoted by $x, y, z$) correspond to the abstract
objects, and
the constants (denoted by $a, b, c$ or italicized strings)
correspond to the atomic values.
Let $s, t$ denote variables or constants, and let
a path (denoted by $p, q$) be a finite, possible empty sequence of
attributes.

\begin{defin}
The terms of the description language $F_L$ are the elements
from $V$ and $A$.
The formulas of the description language ($F_L$-formulas) are
equations, and conjunctions:
\[ ps \doteq qt    \mbox{ and }    \varphi \wedge \psi \]
if $\varphi, \psi$ are formulas, $p, q$ are paths, and $s, t$ are
terms.  The formulas of the following form are called
primitive formulas:
\[ s \doteq t    \mbox{ and }    fs \doteq t . \]
\end{defin}

The description language $F_L$ is interpreted as a special
algebra in \cite{smolka:92}.
However for our purposes it suffices to interpret the formulas
as feature graphs. The formula
$s \doteq t$ is interpreted as: the terms $s$ and $t$ denote the
same node in the feature-graph.
The formula $fs \doteq t$ is interpreted as:
there is an edge with label $f$
from the node denoted by $s$ to the node denoted by $t$
in the feature-graph.

As an example, consider the feature-graph given in Figure~\ref{FSman}.
The following formula describes the feature-graph,
provided that the proper sets $A, V$ and $L$ are given.
\[\begin{array}{c}
   \fn{subject}\,x	\doteq y \:\wedge\:
   \fn{predicate}\,x	\doteq z \:\wedge\:
   \fn{number}\,y	\doteq \fn{number}\,z		\:\wedge\: \\
   \fn{number subject}\,x	\doteq \fv{singular}	\:\wedge\:
   \fn{tense}\,x	\doteq \fv{present}
\end{array}\]

Another familiar, intuitive description is the attribute-value matrix
notation. An attribute-value matrix (AVM) is a set of attribute-value pairs.
The values of the attribute-value pairs are boxlabels, and atomic
values or AVMs, where equal boxlabels denote equal values. The
elements of an AVM are written below one another. The total
set is written between squared brackets.

For instance, the feature-graph given in Figure~\ref{FSman} could be
represented by the following attribute-value matrix.
The box-labels $\rs{1}$ are used to denote that the two attributes
\fn{number} have the same value.
\[\avm{	\fn{subject}  &\avm{\fn{number} &\rs{1}\,\fv{singular}} \\
	\fn{predicate}&\avm{\fn{number} &\rs{1}} \\
	\fn{tense} &\fv{present} } \]
The AVM notation is intuitive because AVMs strongly resemble
feature-graphs. We can view the opening brackets and the
atomic values of an AVM as nodes. The outermost bracket is the
root-node. The attributes of the AVM can be view as edges with the
attribute as their label. The box-labels identify nodes in the
feature-graph.

In this paper we will use both the AVMs and the $F_L$-formulas as a
description language. Because AVMs can be transformed in linear time
into formulas \cite[Section~6]{smolka:92} the use of different
notations should cause no confusion.

\subparagraph{Unification in $F_L$.}
Let $A$ and $B$ be abstract linguistic objects, or feature-graphs,
that are described by the $F_L$-formulas $\varphi$ and $\psi$,
respectively.  The unification of $A$ and $B$ is described by
$F_L$-formula $\varphi \wedge \psi$
if and only if $\varphi \wedge \psi$ describes a
feature-graph.
In the final part of this section
we will present an efficient algorithm that
determines whether an $F_L$-formula describes a feature-graph.
Hence we can view the algorithm as a unification algorithm.

\subparagraph{Unification in AVM.}
Let $A$ and $B$ be abstract linguistic objects, or feature-graphs,
that are described by the AVMs $[F]$ and $[G]$,
respectively.  The unification of $A$ and $B$ is denoted by
$[F] \sqcup [G]$.
The algorithm of the final part of this section can be used
to compute the AVM $[F] \sqcup [G]$ efficiently, in the following way.

First, there is a linear time algorithm that
transforms AVMs into $F_L$ formulas.
Second, the algorithm of the final part of this section can
easily be modified
such that it also outputs the feature-graph that is described by
an $F_L$-formula. Since the modified algorithm will remain
efficient, the feature-graph will be small.
Finally, there is a trivial, linear time, algorithm that transforms
feature-graphs into AVMs.

\paragraph{This feature theory is simple.}
In the remainder of this section we will show that the feature theory
is simple. We will provide an algorithm, called {\sc FeatureGraphSat}, that
determines whether a formula of the description language describes a
feature-graph.
The algorithm is a slight modification of the constraint-solving
algorithm in \cite[Section~5]{smolka:92}.

The algorithm {\sc FeatureGraphSat} can be used to
determine whether two abstract objects can be unified: if the formulas
$\varphi$ and $\psi$ describe abstract objects, then $\varphi \wedge
\psi$ describes their unification if, and only if, the unification
exists. So we may say that the algorithm solves the unification
problem.

The algorithm {\sc FeatureGraphSat} below
determines syntactically whether a formula is
satisfiable in some feature algebra. Because there is a 1--1
correspondence between
satisfiable formulas and feature-graphs, the algorithm determines
whether a formula describes a feature-graph.
The algorithm first transforms any formula by means of syntactic
simplification rules into a normal form. Then
this normal form is checked syntactically in order to see
whether the formula is satisfiable.

The correctness and the complexity of the algorithm {\sc FeatureGraphSat}
follow from \cite[Section~5]{smolka:92}. The function {\sc Transform},
the procedure {\sc Simplify}, the clash-freeness test and
the acyclicity test can all be computed in an amount of time that is
quadratic in the size of the formula $\varphi$.
Hence the algorithm {\sc FeatureGraphSat} takes quadratic time,
and thus shows that the feature theory is indeed simple.

\begin{tabbing}
End\=then \=  \kill
{\sc Algorithm FeatureGraphSat}	\\
\pushtabs Output:M\= \kill
{\sc Input:} \>Formula $\varphi = \bigwedge_i \varphi_i$ from the
		description language. \\
{\sc Output:} \>1) `Yes' if $\varphi$ describes an acyclic
		feature-graph, or \+\\
	2) `No' otherwise.\-\\
\poptabs
{\bf Begin Algorithm}\+\\
Each $\varphi_i$ is of the form $ps \doteq qt$, where $p, q$ are paths,
$s,t$ are terms. \\
{\sc Transform} $\varphi$ into a set of primitive formulas:   \\
		$P = \{\psi_i | \psi_i \,=\, fs\doteq t,
		\mbox{ or } \psi_i \,=\, s\doteq t \}$. \\
{\sc Simplify} the set $P$, yielding set $S$, until no further
		simplification is possible. \\
{\bf If} set $S$ is clash-free and acyclic, \\
{\bf then}\\
	\>{\bf Exit} with answer `Yes', \\
{\bf else} \\
	\>{\bf Exit} with answer `No'. \-\\
{\bf End Algorithm}
\end{tabbing}

\begin{tabbing}
End\=step 00 \=  \kill
{\sc Function Transform}	\\
\pushtabs Output:M\= \kill
{\sc Input:} \>Formula $\varphi = \bigwedge_i \varphi_i$ from the
		description language. \\
{\sc Output:} \>A set of primitive formulas $P = \{\psi_i | \psi_i
\,=\, fs\doteq t, \mbox{ or } \psi_i \,=\, s\doteq t \}$. \\
\poptabs
{\bf Begin Function}\+\\
 $P \:=\: \varphi^\circ$, where \\
{\bf Step 0.} \>$(\bigwedge_i \varphi_i)^\circ :=
		\bigcup_i (\varphi_i)^\circ$ \\
{\bf Step 1.} \>$(ps \doteq qt)^\circ :=
	(ps \doteq y)^\circ \cup (qt \doteq y)^\circ$ , where $y$ is a
	fresh variable \\
{\bf Step 2.} \>$(f_n\ldots f_1 s \doteq y)^\circ :=
	\{ s \doteq y_0, y_n \doteq y \}  \cup \{ f_i y_{i-1} \doteq
	y_i | 1 \leq i \leq n \}$ , where \+\\
	$y_i$ ($1 \leq i \leq n$)
	are fresh variables, and $y$ is a variable
	introduced in step 1. \-\-\\
{\bf End Function}
\end{tabbing}

In the procedure {\sc Simplify} we will use the following
notations. We use $[x/s]P$ to denote the set that is obtained from
$P$ by replacing every occurrence of variable $x$ by term $s$, and
$s \doteq t\,\&\,P$ to denote the set $\{s \doteq t\}\cup P$,
provided that $s \doteq t \not\in P$.

\begin{tabbing}
End\=end \=00 \=  \kill
{\sc Procedure Simplify} (c.f., \cite{smolka:92})	\\
\pushtabs Output:M\= \kill
{\sc Input:} \>Set of primitive formulas $P$.\\
{\sc Output:} \>Simplified set of primitive formulas $S$. \\
\poptabs
{\bf Begin Procedure}\+\\
{\bf Do while} one of the following four simplification rules is
applicable \+\\
   {\bf 1.} \>$(x \doteq s)\,\&\,P \;\rightarrow\;
	(x \doteq s)\,\&\, [x/s]P$ if $x$ occurs in $P$ and $x \neq s$ \\
   {\bf 2.} \>$(a \doteq x)\,\&\,P \;\rightarrow\;
	(x \doteq a)\,\&\,P$ \\
   {\bf 3.} \>$(fx \doteq s) \,\&\, (fx \doteq t) \,\&\, P \;\rightarrow\;
	(fx \doteq s) \,\&\, (s \doteq t) \,\&\, P$ \\
   {\bf 4.} \>$(s \doteq s)\,\&\,P \;\rightarrow\; P$ \- \\
   {\bf End while} \-\\
   {\bf Exit} with the simplified form of set $P$, $S$. \\
{\bf End Procedure}
\end{tabbing}

\begin{lem}
A simplified set of primitive formulas $S$ is clash-free if
\begin{enumerate}
\item $S$ contains no formula $fa \doteq s$, and
\item $S$ contains no formula $a \doteq b$ such that $a \neq b$.
\end{enumerate}
\end{lem}
\begin{proof}
{}From \cite[Proposition~5.4]{smolka:92}.
\end{proof}

\begin{lem}
A simplified set of primitive formulas $S$ is acyclic if, and only
if, $S$ does not contain a sequence of formulas $f_i x_i \doteq
x_{i+1}$ and $f_n x_n \doteq x_1$ ($1 \leq i \leq n$).
\end{lem}
\begin{proof}
By induction on the length of a cycle.
\end{proof}

\section{No upper bound}\label{Snub}
An novice in complexity theory might expect
that a problem is not harder
than the problem's hardest component. However,
combining problems may yield a problem that is harder than each of the
problems when considered separately.
For instance,
\cite{johnson:88} combines context-free grammars with a
simple feature theory similar to the one in Section~\ref{Ssft}.
Of course, both the
satisfiability problem of this feature theory and
the universal recognition problem of context-free grammars are
decidable.
Nevertheless, Johnson shows that the universal recognition problem
of the combination is undecidable in general.
Johnson also proves that this problem is decidable under the restriction that
the context-free grammar does not contain detours. This restriction
is called the `Off-line Parsability Constraint'.

{}From Johnson's work, we see that combining problems may change
the complexity from decidable to undecidable.
We claim that combining problems may change also
the complexity from tractable to intractable.
Hence, even when we confine ourselves to decidable problems,
the complexity of the recognition problem of a unification grammar that
uses some feature theory may be higher than
the complexity of the
satisfiability problem of that feature theory.
The claim shows that even under the Off-line Parsability Constraint
the complexity of the feature theory still does not provide an upper
bound on the complexity of the unification grammar.

In the next section we will present a fixed regular grammar.
Then we combine this regular grammar with the feature theory from
Section~\ref{Ssft} into a unification grammar.
The recognition problem of this unification grammar is decidable,
because the regular grammar does not contain detours.
Finally, we will prove by a reduction from {\sc Satisfiability}
that the recognition problem of this unification grammar is NP-hard,
which proves the claim by example.

\subsection{A fixed regular grammar}\label{Sfrg}
The regular language that we want to recognize is
$(\sharp ((0\cup 1)^* (p \cup\overline{p}))^+  )^*$.
The rules of a regular grammar $G'$ that generates this language are
given in Table~\ref{Tnrg}.

\begin{table}[ht]
\begin{center}
\fbox{
\begin{minipage}{11cm}
\footnotesize
\begin{tabbing}
$SM$\=$\rightarrow\;$	\=$p\;F$   MM\=$\mid p\;F$   MM\=$\mid p\;F$
			MM\=$\mid p\;F$   MM\=$\mid p\;F$ MM\= \kill
$S$\>$\rightarrow$\>$\sharp\;F$	\>
 $\mid \sharp\;T$		\>
  $\mid \epsilon$ \\
$F$\>$\rightarrow$\>$0\;F$	\>
 $\mid 1\;F$			\>
  $\mid p\;F$		\>
   $\mid \overline{p}\;F$	\>
    $\mid p\;T$		\>
     $\mid \overline{p}\;T$ \\
$T$\>$\rightarrow$\>$0\;T$	\>
 $\mid 1\;T$		\>
  $\mid p\;A$		\>
   $\mid \overline{p}\;A$ \\
$A$\>$\rightarrow$\>$B$	\>
 $\mid S$ \\
$B$\>$\rightarrow$\>$0\;B$	\>
 $\mid 1\;B$		\>
  $\mid p\;A$		\>
   $\mid \overline{p}\;A$
\end{tabbing}
\end{minipage}
}
\caption{Nondeterministic regular grammar for
$(\sharp ((0\cup 1)^* (p \cup\overline{p}))^+ )^*$.\label{Tnrg}}
\end{center}
\end{table}
\begin{fact}\label{Flangnrg}
The regular grammar in Table~\ref{Tnrg} generates
the language $(\sharp ((0\cup 1)^* (p \cup\overline{p}))^+)^*$.
\end{fact}

Many other regular grammars could be given for the same language.
However, the one presented, as will be seen later, is sufficient for our
purposes here: that is, the reduction from {\sc Satisfiability}.
Obviously, the recognition problem of fixed regular grammar
takes linear time.

\subsection{Combining a regular grammar and a feature
theory}\label{Srg&ft}
In this section, we will present the unification grammar $G$, which is
a combination of the regular grammar from the previous section and
the feature theory from Section~\ref{Ssft}.
There are multiple formalisms for unification grammars.
Most of these formalisms distinguish two components: a constituent
structure and a feature graph. The two components are related
by a mapping from the nodes in the constituent structure to the nodes
in the feature graph.

Table~\ref{Tgrug} contains the grammar rules of unification
grammar $G$.
The notation for the grammar is similar to \cite{johnson:88}.
The rules of Section~\ref{Sfrg} are annotated with formulas taken
from the feature theory given in Section~\ref{Ssft}.
The set of attributes is $\{\fn{assign},
\fn{new}, \fn{v}, \fn{0}, \fn{1}\}$, the set of atomic values is
$\{\fv{+}, \fv{--}\}$.
The linear rewrite rules describe how constituents are formed.
The formulas indicate how nodes of the feature-graphs
are related to the non-terminals of the rewrite rules.

\begin{table}[ht]
\begin{center}
\fbox{
\begin{minipage}{12cm}
\footnotesize
\begin{tabbing}
$SM$\= $\rightarrow\;$\=
$\fn{assign} \,x_0 \doteq \fn{assign} \,x_1\:\wedge$M\=
$SM$\= $\rightarrow\;$\=
$\fn{assign} \,x_0 \doteq \fn{assign} \,x_1\:\wedge$M\=
$SM$\= $\rightarrow\;$\=
$\fn{assign} \,x_0 \doteq \fn{assign} \,x_1\:\wedge$M\=
\kill
$S$\>$\rightarrow$\>$\sharp\;F$ \>
 $S$\>$\rightarrow$\>$\sharp\;T$ \>
  $S$\>$\rightarrow$\>$\epsilon$ \+\\
	$\fn{assign} \,x_0 \:\doteq\: \fn{assign} \,x_1$
	 \>\>\>$\fn{assign} \,x_0 \:\doteq\: \fn{assign} \,x_1\:\wedge$
	  \>\>\>$\fn{v assign} \,x_0 \:\doteq\: +$ \+\+\+\\
	 $\fn{assign} \,x_0 \:\doteq\: \fn{new} \,x_1$ \-\-\-\-\\
\\
$F$\>$\rightarrow$\>$0\;F$ \>
 $F$\>$\rightarrow$\>$1\;F$ \>
  $F$\>$\rightarrow$\>$p\;F$ \+\\
	$\fn{assign} \,x_0 \:\doteq\: \fn{assign} \,x_1$
	\>\>\>$\fn{assign} \,x_0 \:\doteq\: \fn{assign} \,x_1$
	 \>\>\>$\fn{assign} \,x_0 \:\doteq\: \fn{assign} \,x_1$\-\\
\\
$F$\>$\rightarrow$\>$\overline{p}\;F$ \>
 $F$\>$\rightarrow$\>$p\;T$ \>
  $F$\>$\rightarrow$\>$\overline{p}\;T$ \+\\
	$\fn{assign} \,x_0 \:\doteq\: \fn{assign} \,x_1$
	\>\>\>$\fn{assign} \,x_0 \:\doteq\: \fn{assign} \,x_1\:\wedge$
	 \>\>\>$\fn{assign} \,x_0 \:\doteq\: \fn{assign} \,x_1\:\wedge$
								\+\+\+\\
	$\fn{assign} \,x_0 \:\doteq\: \fn{new} \,x_1$
	 \>\>\>$\fn{assign} \,x_0 \:\doteq\: \fn{new} \,x_1$ \-\-\-\-\\
\\
$T$\>$\rightarrow$\>$0\;T$ \>
 $T$\>$\rightarrow$\>$1\;T$ \+\\
	$\fn{assign} \,x_0 \:\doteq\: \fn{assign} \,x_1\:\wedge$
	\>\>\>$\fn{assign} \,x_0 \:\doteq\: \fn{assign} \,x_1\:\wedge$ \\
	$\fn{new 0} \,x_0 \:\doteq\: \fn{new} \,x_1$
	\>\>\>$\fn{new 1} \,x_0 \:\doteq\: \fn{new} \,x_1$ \-\\
\\
$T$\>$\rightarrow$\>$p\;A$ \>
 $T$\>$\rightarrow$\>$\overline{p}\;A$ \+\\
	$\fn{assign} \,x_0 \:\doteq\: \fn{assign} \,x_1\:\wedge$
	\>\>\>$\fn{assign} \,x_0 \:\doteq\: \fn{assign} \,x_1\:\wedge$ \\
	$\fn{v new} \,x_0 \:\doteq\: +$
	\>\>\>$\fn{v new} \,x_0 \:\doteq\: -$ \-\\
\\
$A$\>$\rightarrow$\>$B$ \>
 $A$\>$\rightarrow$\>$S$ \+\\
	$\fn{assign} \,x_0 \:\doteq\: \fn{assign} \,x_1$
	\>\>\>$\fn{assign} \,x_0 \:\doteq\: \fn{assign} \,x_1$ \-\\
\\
$B$\>$\rightarrow$\>$0\;B$ \>
 $B$\>$\rightarrow$\>$1\;B$ \+\\
	$\fn{assign} \,x_0 \:\doteq\: \fn{assign} \,x_1$
	\>\>\>$\fn{assign} \,x_0 \:\doteq\: \fn{assign} \,x_1$ \-\\
\\
$B$\>$\rightarrow$\>$p\;A$ \>
 $B$\>$\rightarrow$\>$\overline{p}\;A$ \+\\
	$\fn{assign} \,x_0 \:\doteq\: \fn{assign} \,x_1$
	 \>\>\>$\fn{assign} \,x_0 \:\doteq\: \fn{assign} \,x_1$
\end{tabbing}
\caption{The grammar rules of unification
grammar $G$.}\label{Tgrug}
\end{minipage}
}
\end{center}
\end{table}

The second rule in the first line of Table~\ref{Tgrug} will be used
to explain the notation.
The non-terminal on the left-hand side of the rewrite rule is related
to the node denoted by variable $x_0$.
The leftmost non-terminal on the right-hand side of the rewrite rule
is related to the node denoted by variable $x_1$.
The first conjunct of the formula states that the values of the
attributes \fn{assign} is the same for the nodes related to the
non-terminals $S$ and $T$.
The second conjunct requires that the attribute \fn{assign}
of the node related to the non-terminal $S$ has also the same
value as the
attribute \fn{new} of node related to the non-terminal $T$.
We will clarify the use of the grammar by means
of an example.

\begin{ex}
We will show the potential derivation of the string
$w = \sharp 1 0 p \sharp 1 0 \overline{p}$.
On the left of the figures~\ref{FNufirst} and~\ref{FNupot}
the constituent structure trees are given. The non-terminals
are related to nodes in the feature-graphs by undirected arcs.
We present the first steps (figure~\ref{FNufirst}) and the `final'
result (figure~\ref{FNupot}) of the potential derivation. The
reader should check that the feature-graph indeed conforms to the
formulas of the applied rules.

\begin{figure}[ht]
\centerline{
\framebox[15.5cm][l]{
\epsfxsize=14cm
\epsfbox{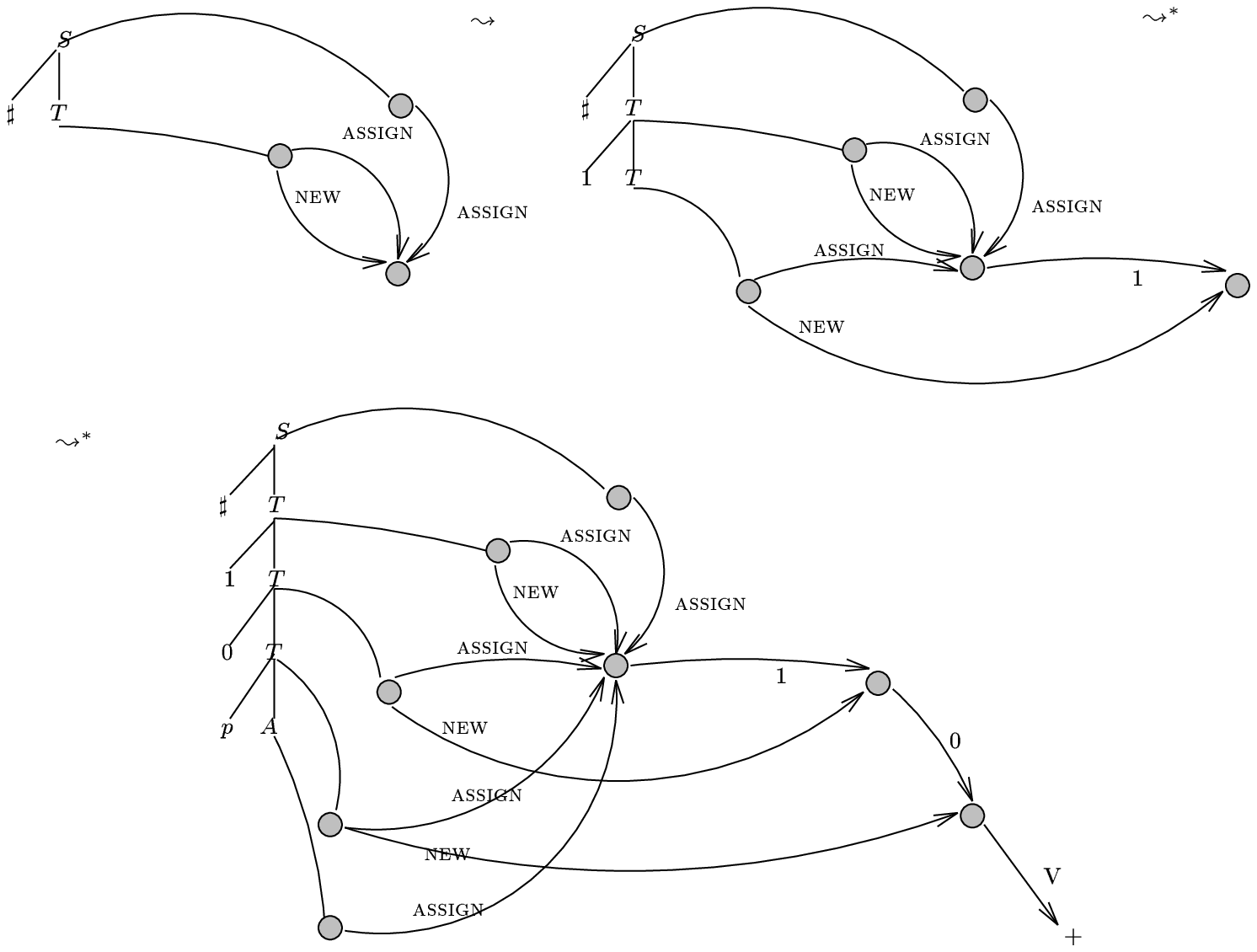}
}
}
\caption{First steps in a potential derivation feature-graph for
$\sharp 1 0 p \sharp 1 0 \overline{p}$.
\label{FNufirst}}
\end{figure}

\begin{figure}[ht]
\centerline{
\framebox[13.0cm][l]{
\epsfxsize=10cm
\epsfbox{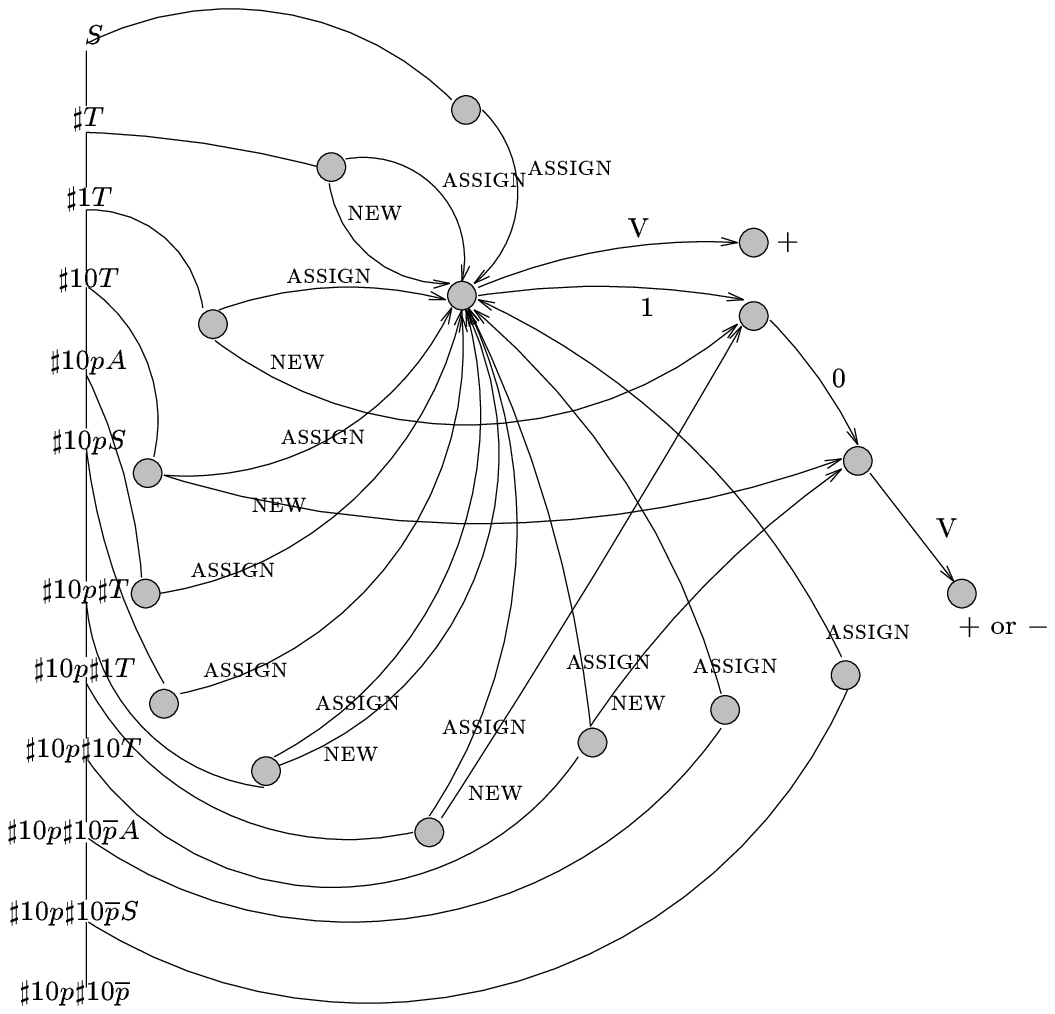}
}
}
\caption{A potential derivation feature-graph for
$\sharp 1 0 p \sharp 1 0 \overline{p}$.
\label{FNupot}}
\end{figure}

The potential feature-graph in figure~\ref{FNupot} shows that
the rightmost
node should have two different atomic values, indicated by $+$ or $-$.
Hence this potential feature graph is not valid.  Consequently, the
derivation given above fails, and the string
$w = \sharp 1 0 p \sharp 1 0 \overline{p}$ cannot be generated.
\end{ex}

The following fact results from fact~\ref{Flangnrg} and the
previous example,
which showed that $w = \sharp 1 0 p \sharp 1 0 \overline{p}$
cannot be generated by $G$.
\begin{fact}
The language recognized by the unification grammar $G$
is a proper subset of the regular language
$(\sharp ((0\cup 1)^* (p \cup\overline{p}))^+  )^*$.
\end{fact}

The following fact will be useful in the proof of
Lemma~\ref{Lfansp}. The fact states that if $S$ derives $w_i\,S$ in
$d$ steps
($S \Rightarrow^d w_i\,S$), then there are two intermediate stages.
First, $S$ derives
$\sharp v^i_1\ldots v^i_{k-1}\,T$ in $a$ steps.
This $T$ derives $v^i_k\,A$ in $b$ steps.
Finally, this $A$ derives $v^i_{k+1}\ldots v^i_n\,S$ in $c$
steps.
\begin{fact}\label{Fderiv}
If $S \Rightarrow^d w_i\,S$, where
$w_i = \sharp v^i_1\ldots v^i_n$, and
$v^i_j \in (0\cup 1)^* (p \cup\overline{p})$, then
there is a $v^i_k = b_1\ldots b_m l$ $(1 \leq k \leq n)$ such that
\[S \;\Rightarrow^a \sharp v^i_1\ldots v^i_{k-1} T
	\;\Rightarrow^b \sharp v^i_1\ldots v^i_{k-1} v^i_k A
	\;\Rightarrow^c \sharp v^i_1\ldots v^i_k v^i_{k+1}\ldots v^i_n S\]
$(d = a + b + c)$
and the feature structure
$[\fn{new} [\fn{$b_1$} \ldots [\fn{$b_m$}\,\alpha]\ldots]]$
is associated with $T$,
where $\alpha = [\fn{v}\,\fv{+}]$ if $l = p$,
and   $\alpha = [\fn{v}\,\fv{--}]$ if $l = \overline{p}$.
\end{fact}

\subsection{The reduction from SAT.}
In the previous section we combined the regular grammar from
Section~\ref{Sfrg} and the feature theory from
Section~\ref{Ssft} into a unification grammar $G$.
Both the recognition problem of this regular grammar,
and the
satisfiability problem of this feature
theory take polynomial time.
However, we will prove that the recognition problem of
the unification grammar $G$ is NP-hard.
Thus the complexity of the
feature theory does not provide an upper bound on the complexity of
the  grammar that used this feature theory.

First, we will give the reduction from the NP-complete problem SAT
to the recognition problem
of $G$. Then we will show that this reduction is
computable in polynomial time
and answer preserving.
Thus we have proven that the recognition problem of
the unification grammar $G$ is NP-hard.

The reduction from SAT to the recognition problem of $G$
maps propositional logical formulas onto strings.
We assume, without loss of generality, that the indices
of the propositional logical variables are in binary representation.
This reduction, $f$, is defined by the following four equations:
\[\begin{array}{rclp{4cm}}
f(\gamma_1\wedge\ldots\wedge\gamma_m) &=
	&\sharp \,f(\gamma_1) \ldots \sharp\,f(\gamma_m)
		&($\gamma_i$ a clause) \\
f(l_1\vee\ldots\vee l_m) &=
	&f(l_1) \ldots f(l_m) &($l_i$ a literal) \\
f(p_i) &=		&i\,p	&($p_i$ a positive literal) \\
f(\overline{p_i}) &=	&i\,\overline{p}
		&($\overline{p_i}$ a negative literal)
\end{array}\]

\begin{fact}\label{Fmaps}
The reduction $f$ maps formula $\varphi$ onto string
$w = f(\varphi) = w_1\ldots w_n$, where
$w_i = \sharp v^i_1\ldots v^i_n$, and
$v^i_j$ is a string of the form $(0\cup 1)^* (p \cup\overline{p})$.
\end{fact}

\begin{lem}\label{Lfinp}
The reduction $f$ is computable in linear time.
\end{lem}
\begin{proof}
By induction on the construction of SAT formulas.
\end{proof}

\begin{lem}\label{Lfansp}
Let $\varphi$ be a propositional logical formula in conjunctive
normalform, and $f$ the reduction stated above.
Formula $\varphi = \gamma_1 \wedge\ldots\wedge \gamma_m$ is a
satisfiable formula if, and only if, string $w = f(\varphi)$ is in the
language generated by $G$.
\end{lem}
\begin{proof}
The proof of this lemma is split in two subproofs. First, we will prove
that if $\varphi$ is satisfiable, then $w$ is in the language generated
by $G$. Second, we will prove that if
$w = f(\varphi)$ is in the language generated by $G$,
then $\varphi$ is satisfiable.

\paragraph{Only if:} let $\varphi$ be a satisfiable formula.
Then there is an assignment $g$ such that
\begin{desc}
\item (1) if $g$ assigns a truth-value to one occurrence of a variable,
then $g$ assigns that truth-value to all occurrences of that variable
in the formula. In other words, $g$ is consistent.
\item (2) $g$ assigns truth to the formula. That is, in each clause,
$g$ assigns truth to some literal.
\end{desc}
We have to show that $w = f(\varphi)$ is generated by $G$. According
to Fact~\ref{Fmaps} $w = w_1\ldots w_m$. This string $w$ is
generated by $G$ if, and only if, the string $w_1\ldots w_m$ is
derived by $S$. Moreover, $S \Rightarrow^* w_1\ldots w_m$ if and
only if $S \Rightarrow^*  w_i S$.
By Fact~\ref{Fderiv}, each derivation $S \Rightarrow^* w_i S$, has the
following intermediate steps:
\[S \;\Rightarrow^* \sharp v^i_1\ldots v^i_{k-1} T
	\;\Rightarrow^* \sharp v^i_1\ldots v^i_{k-1} v^i_k A
	\;\Rightarrow^* \sharp v^i_1\ldots v^i_k v^i_{k+1}\ldots v^i_n S\]
Let us assume that
$S \;\Rightarrow^* \sharp v^i_1\ldots v^i_{k-1} T$, only if the
assignment $g$ assigns truth to the $k$-th literal in the $i$-th
clause of $\varphi$. This $k$-th literal in the $i$-th clause, is either
$p_{b_1\ldots b_l}$ or $\overline{p_{b_1\ldots b_l}}$. In the first
case $g$ assigns truth-value true to variable $p_{b_1\ldots b_l}$, in
the second case $g$ assigns truth-value false to variable
$p_{b_1\ldots b_l}$.
By induction on the number of substrings $w_i$, we will prove that
under the above made
assumption $S$ derives $w_1\ldots w_m$.

\begin{description}
\item[One substring $w_m$:]
Let $S_0 = S$ derive $w_m S$ ($w_m = \sharp v^m_1\ldots v^m_n$),
where $k$ depends on the assignment $g$:
\[  S_0 \;\Rightarrow^* \sharp v^m_1\ldots v^m_{k-1} T
	\;\Rightarrow^* \sharp v^m_1\ldots  v^m_k A
	\;\Rightarrow^* \sharp v^m_1\ldots  v^m_n S
	\;\Rightarrow   \sharp v^m_1\ldots  v^m_n\]
The non-terminal $S$ derives the empty string in one step.
Thus the feature structure associated with $S$ is
$[\fn{assign}\; [\fn{v}\; \fv{+}]]$.
The feature structure associated with $T$ is the unification of
$[\fn{new} [\fn{$b_1$}\ldots [\fn{$b_l$}\,\alpha] \ldots ]]$ and
the feature structure associated with $S$:
\[\avm{  \fn{new} &[\fn{$b_1$}\ldots [\fn{$b_l$}\,\alpha] \ldots ] \\
	 \fn{assign} &[\fn{v}\; \fv{+}] }	\]
where $\alpha = [\fn{v}\,\fv{+}]$ if $v^i_k = b_1\ldots b_l p$,
and $\alpha = [\fn{v}\,\fv{--}]$ if $v^i_k = b_1\ldots b_l\overline{p}$.
The feature structure associated with $S_0$ is
\[
\avm{		&[\fn{$b_1$}\ldots [\fn{$b_l$}\,\alpha] \ldots ] \\
  \fn{assign}	&\sqcup \\
		&[\fn{v}\; \fv{+}] }
=
\avm{ \fn{assign} &\avm{
	\fn{$b_1$}\ldots [\fn{$b_l$}\,\alpha] \ldots \\
	\fn{v}\; \fv{+} }}	\]
None of the unifications fails, and thus $S$ derives $w_m$.

\item[More than one substring $w_i$:]
Let $S_0 = S$ derive $w_i S$ ($w_i = \sharp v^i_1\ldots v^i_n$):
\[  S_0 \;\Rightarrow^* \sharp v^i_1\ldots v^i_{k-1} T
	\;\Rightarrow^* \sharp v^i_1\ldots  v^i_n S \]
By the induction hypothesis, we assume that $S$ derives $w_{i+1}\ldots
w_m$. Moreover, the feature structure associated with $S$ is
$[\fn{assign} [\fn{v}\; \fv{+}]] \sqcup \beta$ = $\beta'$,
where $\beta$ is a feature structure of the form $[\fn{$c_1$}\ldots
[\fn{$c_{l'}$}\,\alpha''] \ldots ]$, or a unification of such feature
structures.
The feature structure associated with $T$ is the unification of
$[\fn{new} [\fn{$b_1$}\ldots [\fn{$b_l$}\,\alpha] \ldots ]]$ and
the feature structure associated with $S$:
\[\avm{ \fn{new} &[\fn{$b_1$}\ldots [\fn{$b_l$}\,\alpha] \ldots ] \\
	\fn{assign} &[\fn{v}\; \fv{+}] \sqcup \beta }
\]

In the case that $v^i_k$ is a prefix of $w_i$ the feature
structure~(\ref{EqfsS}) is associated with $S_0$.
In the other cases, there is an intermediate step
\[S \;\Rightarrow^* \sharp v^i_1\ldots v^i_{k-2} F
	\;\Rightarrow^* \sharp v^i_1\ldots v^i_{k-1} T,	\]
and feature structure~(\ref{EqfsS}) is associated with $F$,
where $\gamma$ is the unification of
$[\fn{$b_1$}\ldots [\fn{$b_l$}\,\alpha] \ldots ]$ and $\beta'$.

\begin{equation}
  \avm{		&[\fn{$b_1$}\ldots [\fn{$b_l$}\,\alpha] \ldots ] \\
  \fn{assign}	&\sqcup \\
		&[\fn{v}\; \fv{+}] \sqcup \beta }	\label{EqfsS}
\end{equation}

In all cases the unification in~(\ref{EqfsS}) fails only if $\beta$
contains $[\fn{$b_1$}\ldots [\fn{$b_l$}\,\alpha'] \ldots ]$, and
$\alpha\sqcup\alpha'$ fails. But, $\alpha\sqcup\alpha'$
fails only if $g$ assigns
both truth-value true and truth-value false to variable
$p_{b_1\ldots b_l}$. Hence
$\alpha\sqcup\alpha'$
would fail only if $g$ would be inconsistent, which $g$ is not.
\end{description}
Hence there is a derivation for string $w = f(\varphi)$
if $\varphi$ is satisfiable.

\paragraph{If:} suppose that $w = f(\varphi)$ is in the language
generated by $G$.
By fact~\ref{Fmaps} $w = w_1\ldots w_m$, where
$w_i = \sharp v^i_1\ldots v^i_m$.
We will prove that for all $i$, there is a $k$ such that
\begin{desc}
\item 1) $S \;\Rightarrow^* \sharp v^i_1\ldots v^i_{k-1} T
	\;\Rightarrow^* \sharp v^i_1\ldots  v^i_n S$
\item 2) the feature structure associated with the non-terminal $S$
	that derives $w$ contains $[\fn{assign}\:[\fn{$b_1$}\ldots
	[\fn{$b_l$}\,\alpha]\ldots ] ]$, where
	$\alpha = [\fn{v}\,\fv{+}]$ if $v^i_k = b_1\ldots b_l p$, and
	$\alpha = [\fn{v}\,\fv{--}]$ if $v^i_k = b_1\ldots b_l\overline{p}$.
\item 3) the feature structure associated with the non-terminal $S$
	that derives $w$ does not contain both
	$[\fn{assign}\:[\fn{$b_1$}\ldots [\fn{$b_l$}
	[\fn{v}\,\fv{+}]]\ldots ] ]$, and
	$[\fn{assign}\:[\fn{$b_1$}\ldots [\fn{$b_l$}
	[\fn{v}\,\fv{--}]]\ldots ] ]$.
\end{desc}
Then the feature structure associated with the non-terminal $S$
that derives $w$ encodes a consistent assignment for $\varphi$ that
makes every clause of $\varphi$ true.

Obviously, $S \Rightarrow^* w$ if, and only if, $S \Rightarrow^* w_i S$.
Hence 1) and 2) follow from fact~\ref{Fderiv}. Because $S$ derives
$w$, the feature structure associated with $S$ does not contain
contradicting information: 3) follows.
This completes the second subproof.
\end{proof}

The previous lemma proves that the reduction $f$ from SAT
to the recognition problem of the unification grammar $G$
is answer preserving. Lemma~\ref{Lfinp} proves that
this reduction $f$ is computable in polynomial time.
Hence these two lemmas together prove that the recognition
problem of the unification grammar $G$ is NP-hard.
\cite{tt:94} show that the complexity result of the
recognition problem for unification grammars
that combine a regular grammar and
the feature theory from Section~\ref{Ssft} is strengthened.
An additional NP upper bound is proven
for an arbitrary string and grammar,
which results in an
NP-complete recognition problem.

\begin{lem}
Let $w$ be any string and $G$ be any
unification grammar that combines a regular grammar and
the feature theory from Section~\ref{Ssft}.
Then the recognition problem for $w$ and $G$ is NP-complete.
\end{lem}
\begin{proof}
An NP-hard lower bound is proven above. An NP upper bound
is proven when we can guess a solution, and check
that solution in polynomial time.
The NP upper bound is proven as follows.

Given a string $w$ and a grammar $G$, we can guess a sequence of
$O(|w|)$ rules that encode the derivation for $w$. The guessed
rules describe a constituent structure tree and a set of formulas.
First, we must check that the constituent structure tree described
by the rules has yield $w$. Second, we have
to check that the set of formulas describes some feature-graph.

The first check is trivial. The second check is performed by the
algorithm {\sc FeatureGraphSat} from Section~\ref{Ssft}. Clearly,
both checks only take polynomial time.
\end{proof}

\section{On lower bounds}\label{Snlb}
The previous section shows the complexity of a feature
theory does not provide an upper bound for the complexity
of a unification grammar that uses this feature theory.
The question that arises is whether the complexity of a feature
theory provides a lower bound for the complexity
of such a unification grammar.

In general, it seems that the complexity of the combination of
two problems is at least as hard as the complexity of these
two problems in isolation. So one would be tempted to answer
the question
above in the affirmative. However, if a problem $A$ contains
information about solutions for a problem $B$, and vice versa,
then the combination of $A$ and $B$ may have lower complexity than $A$
and $B$ in isolation.  For instance, let problem $A$ be the complement
of problem $B$. Then the combinations `$A$ or $B$' and `$A$ and $B$'
have the trivial solutions `always answer yes' and `always answer no',
respectively.

To be more specific, in the case of unification grammars,
there seem to be easy reductions from the unification problem
of a feature theory to the recognition problem of arbitrary
unification grammars that use this feature theory.
In some specific situations, however, these reductions do not exist.
Below, we will present some examples of situations in which the
feature theory does not provide a lower bound for the
recognition problem.

\begin{ex}
\begin{itemize}
\item The feature theory does not provide a lower bound if
	the complexity of the recognition problem of the
	grammar component provides a lower bound for the complexity of
	the recognition problem of the unification grammar.
	Consider for instance the class of grammars that generate
	a finite language.
	The combination of a feature theory with a grammar from this
	class yields a unification grammar that generates a finite
	language. Obviously, the recognition problem of this
	unification grammar
	does not depend on the unification problem of the feature
	theory. Hence the lower bound complexity of this class of
	unification grammars is not provided by the complexity of the
	feature theory.
\item The feature theory does not provide a lower bound if the
	unification grammar uses only a fragment of the feature theory.
	This happens when the unification grammar formalism
	restricts the unification. For instance, the unification grammar
	formalism may demand that feature structures are unified at the
	outermost attributes. This demand implies that the size of the
	feature structures that appear in the fixed unification grammar is
	bounded. Consequently, there have to be feature structures in
	the feature theory that cannot be encoded by the unification
	grammar.

	One may object that the obligatory
	unification at the outermost attribute should be incorporated
	in the formalization of the feature theory. Thus reducing
	the complexity of the unification problem of the feature theory.
	However, there is no predefined way to construct unification
	grammars from a feature theory and a grammar component. So,
	there may be many blurred restrictions on the unification.
	These blurred restrictions are the cause that the formalization
	of the feature theory
	may be too expressive and that the unification grammar uses
	only a fragment of the feature theory.
\end{itemize}
\end{ex}

The two examples show that not in all
situations the complexity of the unification problem of the
feature theory provides a lower bound for the complexity of
the recognition problem of the unification grammar. In some special
cases the complexity of the unification grammar may be lower than
the complexity of the feature theory.
Hence care has to be taken for
drawing overhasty conclusions about the lower bound complexity of
the unification grammar from the complexity of the feature theory.

\section{Conclusions}\label{Sconc}
In this paper, we have assessed the complexity results of
formalizations that intend to describe feature theories in
computational linguistics.
These formalizations do not take the constituent structure
component of unification grammars into account. As a result,
the complexity of the unification problem of
feature theories does not provide an upper bound, and need not provide
a lower bound, for the complexity of the recognition problem of
unification grammars using these theories.

Thus the complexity results that have been achieved in
the formalisms of feature theories
are not immediately relevant for unification
grammars used in computational linguistics.
Complexity analyses will only contribute to computational linguistics
if the analyzed formalizations are connected closely with actual
unification grammars.
Therefore, we argue for formalisms that describe
unification grammars as a whole instead of bare feature theories.

\newcommand{\etalchar}[1]{$^{#1}$}

\end{document}